# Mimic: An adaptive algorithm for multivariate time series classification

**Yuhui Wang, Diane J. Cook**
Washington State University
mark166.wang@wsu.edu, djcook@wsu.edu

## Abstract

Time series data are valuable but are often inscrutable. Gaining trust in time series classifiers for finance, healthcare, and other critical applications may rely on creating interpretable models. Researchers have previously been forced to decide between interpretable methods that lack predictive power and deep learning methods that lack transparency. In this paper, we propose a novel Mimic algorithm that retains the predictive accuracy of the strongest classifiers while introducing interpretability. Mimic mirrors the learning method of an existing multivariate time series classifier while simultaneously producing a visual representation that enhances user understanding of the learned model. Experiments on 26 time series datasets support Mimic's ability to imitate a variety of time series classifiers visually and accurately.

## 1 Introduction

A physician making a critical diagnosis will value insights provided from time series sensor data, but multivariate time series data often defy human interpretation. Time series classification (TSC) builds predictive models from a time-ordered set of feature values. The physician will only utilize methods with a history of strong predictive performance. To ensure the patient's health, however, the doctor will also want to understand the reason for the classifier's prediction. This balance between classification accuracy and interpretability has become crucial for many high-stakes applications, including finance, security, and weather, as well as health care.

TSC has made steady progress over the years. Research has traditionally focused on classification of univariate data. This focus departs from reality because multivariate TSC (MTSC) problems abound in most areas of decision making. With the growing need to solve the MTSC problem, deep learning methods have been harnessed to improve predictive accuracy. However, these black-box methods sacrifice model transparency and interpretability.

In contrast with deep networks, shapelet-based classifiers offer a strategy for interpreting time series models. These algorithms discover repeated patterns in a time series that are most representative of a target class. The patterns can be visualized in a low-dimensional space, providing some understanding of the learned model. This strategy comes at a possible cost of decreased classification accuracy, particularly for complex, multivariate data.

An interpretable time series classifier should offer the predictive performance of powerful black-box algorithms and the visualization capability of shapelet classifiers. To address this need, we proposed a novel algorithm, called Mimic. Mimic generates a set of 2D shapelets that imitate any model learned by a time series classifier. Mimic is an adaptive algorithm that can be paired with multiple types of time series classifiers. This algorithm in essence "mimics" the performance of existing time series classifiers.

Our contributions are summarized as follows: (1) We proposed a high accuracy method without sacrificing transparency. (2) The method we propose can adapt to many classifiers, offering a novel way to understand 'black-box' classifiers. (3) Based on our studies, we have shown the method is applicable and effective for multiple types of datasets and classifiers.

## 2 Related Work

Recent work on multivariate time series classification (Karlsson, Papapetrou, and Boström 2016; Baydogan and Runger 2015; Schäfer and Leser 2018; Wistuba, Grabocka, and Schmidt-Thieme 2015; Bagnall et al. 2018) follows several directions. These directions include time series shapelet discovery and classifier design based on dictionaries, distance measures, intervals, and shapelet frequencies. Shapelets (Ye and Keogh 2009) are subsequences of time series data that maximally represent a class. Wistuba et al. (Wistuba, Grabocka, and Schmidt-Thieme 2015) introduced Ultra Fast Shapelets (UFS), an approach to efficiently select such representative patterns from multivariate time series data. A challenge that accompanies shapelet discovery is the large computational cost associated with finding discriminative subsequences, which becomes even more demanding when processig multivariate time series. UFS solves this by extracting random shapelets. Even though UFS solves the computational problem, UFS performance on multivariate time series data is not stable.

In an effort to combine shapelet discovery with traditional supervised learning, Karlsson et al. (2016) introduced Generalized Random Shapelet Forests (gRSF) to generate decision trees from randomly-selected shapelets. In this approach, each node in the decision tree represents one of the random shapelets. An appropriate branch is traversed based on whether or not the new time series contains the corresponding shapelet. Employing a similar strategy, the Generalized Random Forest (gRFS) represents an ensemble of weak learners in which p generalized trees are grown. As with gRSF, each node in the trees contains a single shapelet. To introduce variability amongst the constituent classifiers, gRFS adds bagging. Experiments show that gRSF outperforms other shapelet-based algorithms, including UFS.

Aaron and Anthony (2017) proposed a shapelet transform for multivariate time series classification (STC) to convert the original time-series data to a shapelet set. During this conversion, only the data sequences that correspond to discovered shapelets are retained by replacing data with the shapelet together with the corresponding time values where it occurs. Their work extends shapelet-based classifiers from decision trees to other classification methods.

The second type of approach is dictionary-based algorithms which utilize bag-of-patterns learning strategies. Dictionary-based algorithms move a sliding window unidirectionally over the time series to extract subsequences. Such approaches transform each subsequence into a word using methods such as Piecewise Aggregate Approximation and Symbolic Aggregate approXimation(Sun et al. 2014). These processes transform each time series into a bag, or collection, of words. Based on this bag, Symbolic Aggregate approXimation derives the frequencies of each word for each time series and uses these values to describe the data. As an example of this type of approach, Baydogan and Runger (2015) introduced an approach named Symbolic Representation for Multivariate Time series (SMTS). SMTS trains a random forest from the multivariate time series "bag" to partition the data into leaf nodes, each represented by a word to form a codebook. A second random forest is trained using the words in this codebook to classify the multivariate time series. Recently, Schafer and Leser (2018) introduced an approach named WEASEL-MUSE, which uses the bag of SFA (Symbolic Fourier Approximation) symbols to classify multivariate time series.

Originally a univariate time series classifier, Word Extraction for Time Series Classification, WEASEL (Schäfer and Leser 2017) was extended to include the Multivariate Unsupervised Symbols and Derivatives (MUSE) (Schäfer and Leser 2018), thus enhancing the algorithm to handle multivariate time-series classification (MTSC). Here, words in the form of unigrams and bigrams are extracted for all series and dimensions using a sliding window for a range of window lengths.

The other type of approach includes distance-based models, frequency-based models, and interval-based models. K nearest neigbours(with dynamic time wrapping) (KNN-DTW) (Bagnall et al. 2016) is a distance-based model for time series classification. KNN-DTW compare each object with all the other objects in the training sets to make the classification. The limitations of the KNN-DTW include (1) poor performance with noisy data (2) poor time complexity and space complexity. The time series forest (TSF) classifier (Deng et al. 2013) is an interval-based classifier. TSF splits the time series into the random interval with the random start position and random lengths. Then TSF extracts summary statistics features for each interval and trains a decision tree based on the extracted features. TSF repeats this process until it reaches the required number of trees. Random interval spectral ensemble (RISE) (Flynn, Large, and Bagnall 2019) is a frequency-based model. RISE is similar to TSF, except RISE uses a single time interval for each tree, and RISE uses spectral features.

In addition to these methods, deep learning-based techniques (Karim et al. 2019; Zheng et al. 2014;

Zhang et al. 2020; Ismail Fawaz et al. 2020; Szegedy et al. 2015) achieved promising performance in multivariate time series classification tasks. These models often use an LSTM layer and a stacked CNN layer to extract features from the time series. A softmax layer is then added to predict the class value. Recent approaches employ a CNN to capture relationships between the features and an LSTM to capture dependencies within the time interval. Zheng et al. (Zheng et al. 2014) introduced a model named Multi-Channel Deep Convolutional Neural Network (MCDCNN), for which each univariate time series is passed through a separate CNN layer. The proposed architecture represents a traditional deep CNN with one modification for multivariate time-series (MTS) data: the convolutions are applied independently (in parallel) on each dimension (or channel) of the input MTS. The outputs of all univariate time series are concatenated and passed through a softmax layer to predict the label. While this method is efficient, a limitation of the approach is that each dimension is handled independently; thus, the learned concept does not identify nor utilize relationships between the channels. In contrast, Karim et al. (Karim et al. 2019) recently proposed a model consisting of an LSTM layer and stacked CNN layer along with a Squeeze-and-Excitation block to generate latent features. The Squeeze-and-Excitation Block is an architectural unit designed to improve the representational power of a network by enabling it to perform dynamic channel-wise feature recalibration.

In contrast to conventional approaches, deep learning-based methods can learn latent features by training convolutional or recurrent networks with large-scale labeled data. As an example, Residual network (ResNet)(Wang, Yan, and Oates 2017) was first applied to time series classification by Wang et al. This network contains three consecutive blocks, each comprised of three convolutional layers, which are connected by residual 'shortcut' connections that tie the input of each block to its corresponding output.

In other work that utilizes deep architectures, Zhang et al. (Zhang et al. 2020) proposed a time series attentional prototype network (TapNet). Time series attentional prototype networks are aimed at tackling problems in the multivariate domain. The TapNet architecture draws on the strengths of both traditional and deep learning approaches. Zhang et al. note that deep learning approaches excel at learning low-dimensional features without the need for embedded domain knowledge, whereas traditional approaches such as a 1NN-DTW (Kate 2016) work well on comparatively small datasets. TapNet combines these advantages to produce a network architecture that can be broken down into three distinct modules: Random Dimension Permutation, Multivariate Time Series Encoding, and Attentional Prototype Learning.

Recently, Fawaz et al. (Ismail Fawaz et al. 2020) introduced InceptionTime (IT). InceptionTime achieves high accuracy through a combination of building on ResNet to incorporate Inception modules (Szegedy et al. 2015) and ensembling over five multiple random-initial-weight instantiations of the network for greater stability. A single network out of the ensemble is composed of two blocks of three Inception modules each, as opposed to the three blocks of three traditional convolutional layers in ResNet. These blocks maintain residual connections and are followed by global average pooling and softmax layers as before. InceptionTime outperforms other methods, such as ResNet, gRFS and TapNet, in multivariate time-series classification tasks.

The approaches we describe here are accompanied by different strengths and weaknesses. The shapelet-based algorithms exhibit weaker performance on MTSC tasks comparing to bag-of-pattern and deep learning approaches. On the other hand, the bag-of-patterns and deep learning approaches lack transparency, which limits understanding of how the model makes the classification. To address this limitation while retaining the predictive power, we proposed our MimicShape algorithm. The proposed method can Mimic the performance of these "black-box" algorithms while providing a way to understand how the model is generating classifications.

## 3 Mimic

The Mimic method explores the region impact pre-trained classifiers' classification. To simulate the pre-trained classifiers, the Mimic finds similar patterns from these high-impact regions to make the classifications.

## 3.1 Definitions

We define a multivariate time-series as a matrix $\chi$, where $\chi$ is delineated by the number of time steps, T, and the number of data features V, thus $\chi$ is a matrix with size V x T. We let L represent the set of possible class labels corresponding to $\chi$. The goal of a MTSC is to learn the function $f: \chi \rightarrow y$. This classification function is learned by a pretrained classifier that Mimic will emulate. We assume that f has been supplied from another source and will vary depending on the type of classifier that we are mimicking. The Mimic algorithm also requires that the classifier generate a probability distribution $Pr(\chi, y)$ over possible class labels for sample $\chi$.

Mimic employs a random mask to identify the most predictive elements of the input data. We define random mask M as a set of binary filters in which each filter has dimensions that are the same as the input vector $\chi$. Similarly, we define an importance map I as a matrix of real values between 0 and 1, in which the dimensions of the matrix are the same as the input vector $\chi$. Finally, we define a MimicShape MS as a set of subsequences from the input time-series data. These subsequences are identified by the Mimic algorithm and provided as a visual interpretation of the learned classification model. MS represents visualizations of the time-series sequences that most greatly helped the target classifier make accurate predictions.

To obtain the best performance for Mimic, we first normalize the input data. The traditional normalization method is max-min normalization. We introduce a refinement of this method which re-scales feature values to be distributed between 0 and 1. The traditionally max-min normalization method rescale the feature value $X_{norm\_traditional} \in [0,1]$. However, to accommodate the mask filter operation, all normalized values should be strictly greater than 0. Thus, we modify max-min normalization to follow Equation 1.

$$X_{norm} = \frac{x - \min(x) + 1}{\max(x) - \min(x) + 1} \quad (1)$$

As a result, all normalized values are in the range (0,1].

## 3.2. Architecture

The architecture for our multidimensional Mimic detection model contains four components. These components are a random mask, a pre-trained classifier, a map constructor, and the MimicShape generator. The Mimic architecture is illustrated in Figures 1 and 2. Figure 1 shows the process of constructing the importance map from the input time-series data, while Figure 2 demonstrates the process of finding the MimicShape from the data based on the importance map.

**Algorithm 1** Mimic
1  **Function** Mimic ( $\chi, y, L, Pr$ )
   **Input:**
     X: the normalized time-series data
     y: the corresponding labels
     L: the set of all labels $L$
     Pr: the probability for the given input and label based on the pretrained classifier
   **Output:** MimicShape MS
2     Masks ← RandomMask($\chi$)
3     **For** Mask **in** Masks
4        MaskData ← $\chi$ × Mask
5        ConfidenceScore[Mask,y] ← $Pr(\chi, y)$
6     **For** $l$ **in** $L$
7        ImportanceMap[ $l$ ] ← MapGenerator(ConfidenceScore, $l$ )
8        MS[ $l$ ] ← MimicShapeGenerator(ValueSequence[ $l$ ], $\chi$)
9     **Return** MS
10    **End**

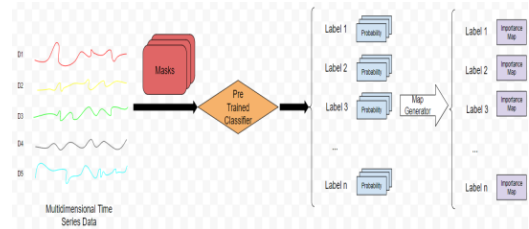

Figure 1. The process of generating the importance maps.

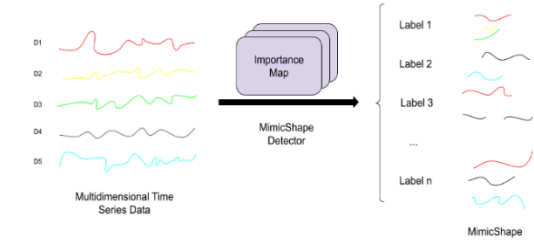

Figure 2. The process of finding MimicShape from the time-series data based on the importance maps.

The whole process of the MimicShape detection is described as follows: First, the time-series data will be masked before being given to the classifier. Then, the classifier will make a prediction based on the masked data. The first two steps are repeated until the model tries all masks generated from the random-mask component.

The importance map generator works by using the classifier's generated probability distribution, $Pr$. The importance map generator produces an importance map by computing a weighted sum of random masks. Each importance map only corresponds to one label. Given the importance map for each label, the MimicShape can be detected by the MimicShape generator. The MimicShape generator captures repeated patterns from the time-series data, and the importance map will constrain the position and length of the MimicShape. The pseudocode for this process is described in Algorithm 1.

### 3.3. Importance Maps

Importance map is a tool that helps Mimic to finding how the target classifier making decisions. It can highlight the influence coordinates when the target classifier gave a classification. By giving these coordinates, we can explore the learning concept of the target classifier. Generating an importance map is relied on random masks. Random masks blocking some coordinates of the input data can help define the influence for each coordinate.

Let random mask $M: \Lambda \rightarrow \{0,1\}$ be a random binary mask with the distribution $\mathcal{U}$. Consider the random variable $Pr(\chi \odot M)$ Where $\odot$ denotes element-wise multiplication. First, the input vector is masked by preserving only a subset of coordinates. An example is shown in Figure 3. Then, the confidence score for the masked data is computed by the target classifier.

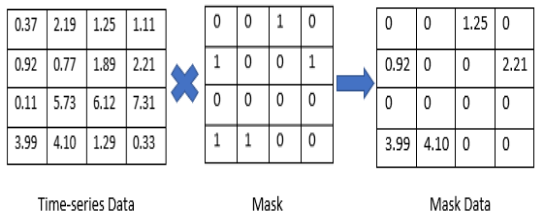

Figure 3. An example of applying a mask to the time-series data.

Next, we define the *importance* of coordinates $\beta$ as the expected score over all possible masks $M$ conditioned on the event that coordinate $\beta$ is observed, denoted as $M(\beta) = 1$:

$$I_{\chi,f}(\beta) = E_M[Pr(\chi \odot M)|M(\beta) = 1] \quad (2)$$

The intuition behind this is that $Pr(\chi \odot M)$ is high when coordinates preserved by mask $M$ are essential. The non-essential coordinates do not have a big impact on the classifier making the decision. Thus, the non-essential coordinates that have been masked will not have a huge influence, which decreases $Pr(\chi \odot M)$.

Equation (2) can then be rewritten as a summation over mask $m \in M$:

$$I_{\chi,f}(\beta) = \sum_m Pr(\chi \odot M)P[M = m|M(\beta) = 1] \quad (3)$$
$$= \frac{1}{P[M(\beta) = 1]} \sum_m Pr(\chi \odot M)P[M = m, M(\beta) = 1]$$

$$P[M = m, M(\beta) = 1] = \begin{cases} 0, & if\ M(\beta) = 0 \\ P(M = m), & if\ M(\beta) = 1 \end{cases} \quad (4)$$
$$= m(\beta)P(M = m)$$

Substituting $P[M = m, M(\beta) = 1]$ from (4) in (3),

$$I_{\chi,f}(\beta) = \frac{1}{P[M(\beta)=1]} \sum_m Pr(\chi \odot M) \cdot m(\beta) \cdot P(M = m) \quad (5)$$

This formula can be written in matrix notation, combined with the fact that $P[M(\beta) = 1] = E[M(\beta)]$:

$$I_{\chi,f} = \frac{1}{E[M]} \sum_m Pr(\chi \odot M) \cdot m \cdot P(M = m) \quad (6)$$

Thus, the importance map can be computed as a weighted sum of random masks, where weights are the probability scores adjusted for the distribution of the random masks.

We propose to generate importance maps by empirically estimating the sum in Equation 6 using Monte Carlo sampling. To produce an importance map, we sample a set of masks $\{M_1, M_2 \cdots M_N\}$ according to $\mathcal{U}$ and probe the model by running it on masked time-series data $\chi \odot M, i = 1, \cdots, N$. Then, we compute the weighted average of the masks. Here, the weights represent the Probability $Pr(\chi \odot M)$. The importance map is then normalized using the expectation of M:

$$I_{\chi,f}(\beta) \stackrel{MC}{\approx} \frac{1}{E(M) \cdot N} \sum_{i=1}^N Pr(\chi \odot M) \cdot M_i(\beta) \quad (7)$$

After we obtain the importance map $I$, we can generate the MimicShape. To find the best subsequence that corresponds to the classifier's learned model, we need to make some modifications to the original importance map.

From the original importance map, we now want to create a new version of the map with only binary [0,1] values. We want to capture all possible patterns in the time-series data which have an influence on the classification. To accomplish this, the new map, $I_{mimic}$ contains 0 values wherever the original map contained 0 values. The new map replaces non-zero values with 1.

The MimicShape detector applies the dynamic time wrapping (DTW) algorithm with the input time-series data $\chi_I$ constrain by the importance map $I_{mimic}$.

$$\chi_I = \chi \cdot I_{mimic} \quad (8)$$

Since the importance map $I_{mimic}$ is a binary map that only has binary [0,1] values, the constrained data $\chi_I$ will be segmented by any 0 value in the $I_{mimic}$. Therefore, we consider each dimension separately. Also, we do not want the detector to find a sequence of 0 values as a pattern. So the detector will separate the sequence into some subsequences which not contain 0 values. The detector finds patterns in these subsequences.

## 4. Experimental Results

We hypothesize that a black-box classifier will yield better predictive performance than the shapelet-based algorithm for some multivariate datasets. We further postulate that our Mimic algorithm, which is interpretable, can simulate the performance of the target black-box classifier without significant degradation in performance. To evaluate Mimic, we conduct experiments on the UEA MTSC archive. For the 2018 release, the UEA MTSC archive contained 30 multivariate time-series datasets. In four of the datasets, instances did not have uniform lengths. As a result, we restrict our attention to the 26 datasets with equal-length time-series. A summary of the main characteristics for each dataset is provided in Table 1.

| Code | Name | Train size | Test size | Dims | Length | Classes |
|---|---|---|---|---|---|---|
| AWR | ArticularyWordRecognition | 275 | 300 | 9 | 144 | 25 |
| AF | AtrialFibrillation | 15 | 15 | 2 | 640 | 3 |
| BM | BasicMotions | 40 | 40 | 6 | 100 | 4 |
| CR | Cricket | 108 | 72 | 6 | 1197 | 12 |
| DDG | DuckDuckGeese | 50 | 50 | 1345 | 270 | 5 |
| EW | EigenWorms | 128 | 131 | 6 | 17,984 | 5 |
| EP | Epilepsy | 137 | 138 | 3 | 206 | 4 |
| EC | EthanolConcentration | 261 | 263 | 3 | 1751 | 4 |
| ER | ERing | 30 | 270 | 4 | 65 | 6 |
| FD | FaceDetection | 5890 | 3524 | 144 | 62 | 2 |
| FM | FingerMovements | 316 | 100 | 28 | 50 | 2 |
| HMD | HandMovementDirection | 160 | 74 | 10 | 400 | 4 |
| HW | Handwriting | 150 | 850 | 3 | 152 | 26 |
| HB | Heartbeat | 204 | 205 | 61 | 405 | 2 |
| LIB | Libras | 180 | 180 | 2 | 45 | 15 |
| LSST | LSST | 2459 | 2466 | 6 | 36 | 14 |
| MI | MotorImagery | 278 | 100 | 64 | 3000 | 2 |
| NATO | NATOPS | 180 | 180 | 24 | 51 | 6 |
| PD | PenDigits | 7494 | 3498 | 2 | 8 | 10 |
| PEMS | PEMS-SF | 267 | 173 | 963 | 144 | 7 |
| PS | PhonemeSpectra | 3315 | 3353 | 11 | 217 | 39 |
| RS | RacketSports | 151 | 152 | 6 | 30 | 4 |
| SRS1 | SelfRegulationSCP1 | 268 | 293 | 6 | 896 | 2 |
| SRS2 | SelfRegulationSCP2 | 200 | 180 | 7 | 1152 | 2 |
| SWJ | StandWalkJump | 12 | 15 | 4 | 2500 | 3 |
| UW | UWaveGestureLibrary | 120 | 320 | 3 | 315 | 8 |

Table 1. Summary of the 26 UEA datasets used in experimentation.

Because we offer an alternative to traditional shapelet classifiers, we compare Mimic with two traditional shapelet algorithms. Because we introduce an approach that can be applicable for other types of classifiers as well, we also combine Mimic with three non-shapelet classifiers. In each case, we evaluate the ability of Mimic to generate interpretable models that emulate the underlying classifier. The shapelet-based classifiers we pick are Shapelet Transform Classifier (STC) and generalized Random Shapelet Forest (gRFS), which are state-of-the-art for shapelet-based multivariate classifiers. The STC select best k shapelets and transform data to set of shapelets. The gRFS use shapelets to generate a random forest for classification. These two methods have demonstrate some of the strongest performances among shapelet-based approaches. We then select two neural network classifiers, Multivariate LSTM-

| | STC (%) | gRFS | FCM-LSTM | Mimic FCM-LSTM | IT | Mimic IT | Decision Tree | Mimic Decision Tree | Rocket | Mimic Rocket | HIVE-COTE | Mimic Hive COTE | RISE | Mimic RISE |
|---|---|---|---|---|---|---|---|---|---|---|---|---|---|---|
| AWR | 97.51 | 98.21 | 98.13 | 98.17 | **99.56** | 99.46 | 47.32 | 94.94* | 99.56 | 99.43 | 97.99 | 97.56 | 95.73 | 94.57 |
| AF | **31.78** | 27.53 | 26.12 | 25.10 | 22.00 | 22.00 | 25.19 | 22.44* | 24.89 | 24.31 | 29.33 | 28.18* | 24.44 | 23.90 |
| BM | 97.92 | **100.00** | **100.00** | 99.75 | **100.00** | 100.00 | 33.10 | 99.20* | 99.00 | 99.75 | 100.00 | 99.75 | 100.00 | 98.75 |
| CR | 98.94 | 97.41 | 98.38 | 98.20 | 99.44 | 99.34 | 56.11 | 89.33* | **100.00** | 99.21 | 99.26 | 99.21 | 97.78 | 97.53 |
| DDG | 43.47 | 44.47 | 56.00 | 56.00 | **63.47** | 62.47 | 37.92 | 52.67* | 46.13 | 45.31 | 47.60 | 45.57* | 50.80 | 48.12* |
| EW | 74.68 | 83.00 | 45.17 | 44.29 | 44.37 | 43.27 | 42.17 | 77.30* | **86.28** | 85.42 | 78.17 | 75.3* | 81.93 | 81.53 |
| EP | 98.74 | 96.01 | 98.38 | 98.30 | 98.68 | 98.48 | 78.93 | 91.02* | 99.08 | 96.73* | **100.00** | 98.22 | 99.86 | 99.67 |
| EC | **82.36** | 34.06 | 28.99 | 28.33 | 27.92 | 26.72 | 33.19 | 29.57* | 44.68 | 44.52 | 80.68 | 76.37 | 49.16 | 48.79 |
| ER | 84.28 | 91.98 | 95.65 | 94.77* | 92.10 | 92.10* | 52.67 | 90.88* | **98.05** | 97.19 | 94.26 | 92.19 | 82.44 | 81.35 |
| FD | 69.76 | 55.36 | 68.89 | 66.39* | **77.24** | 76.21* | 53.12 | 63.13* | 69.42 | 68.31* | 69.17 | 66.30* | 51.17 | 51.47 |
| FM | 53.40 | 54.43 | 53.90 | 52.11* | **56.13** | 55.83 | 54.30 | 45.20* | 55.27 | 54.12 | 53.77 | 52.11* | 52.10 | 51.98 |
| HMD | 43.47 | 32.07 | **52.21** | 52.17 | 42.39 | 42.39 | 29.77 | 30.72* | 44.59 | 38.73* | 37.79 | 37.52 | 28.24 | 27.79 |
| HW | 28.77 | 36.96 | 35.13 | 34.10 | **65.74** | 63.96 | 16.95 | 59.54* | 56.67 | 54.11 | 50.41 | 48.72* | 18.27 | 23.70* |
| HB | 72.15 | 74.89 | **76.52** | 75.44* | 73.20 | 72.50* | 41.36 | 68.07* | 71.76 | 66.75* | 72.18 | 71.02* | 73.22 | 73.19 |
| LIB | 84.46 | 75.56 | 83.67 | 83.47 | 88.72 | 87.89 | 57.93 | 83.43* | **90.61** | 89.37 | 90.28 | 87.39* | 81.67 | 81.25 |
| LSST | 57.82 | 58.15 | 56.17 | 56.10 | 33.97 | 33.87 | 33.91 | 56.19* | **63.15** | 62.87 | 53.84 | 52.18* | 50.58 | 50.43 |
| MI | 50.83 | 51.87 | 51.80 | 51.80 | 51.17 | 51.07 | 37.10 | 49.33* | **53.13** | 53.01 | 52.17 | 51.78 | 49.83 | 49.62 |
| NATO | 84.35 | 82.37 | 84.41 | 84.33 | **96.63** | 95.43 | 26.12 | 81.08* | 88.54 | 86.79 | 82.85 | 82.77 | 80.59 | 80.10 |
| PD | 97.70 | 96.12 | 98.97 | 98.97 | **99.68** | 99.78 | 41.57 | 97.22* | 99.56 | 99.25 | 97.19 | 96.83 | 87.47 | 87.33 |
| PEMS | 98.40 | 91.27 | **99.85** | 98.39 | 82.83 | 82.93 | 57.12 | 78.27* | 85.63 | 84.32 | 97.98 | 95.92 | 98.98 | 98.75 |
| PS | 30.62 | 22.71 | 26.56 | 25.44 | **36.74** | 35.74 | 22.37 | 14.37* | 28.35 | 27.69 | 32.87 | 32.24 | 26.78 | 25.92* |
| RS | 88.09 | 87.79 | 89.30 | 88.20* | 91.69 | 90.59* | 62.39 | 85.39* | **92.79** | 91.35* | 90.64 | 90.47 | 84.17 | 83.86 |
| SRS1 | 54.73 | 79.74 | 85.94 | 84.17 | 84.69 | 82.69 | 20.38 | 79.64* | **86.55** | 86.10 | 86.02 | 85.88 | 73.17 | 73.12 |
| SRS2 | 51.63 | 48.69 | 48.87 | 48.80 | **52.04** | 51.89 | 42.71 | 51.43* | 51.35 | 50.88 | 51.67 | 51.32 | 50.28 | 49.87 |
| SWJ | 44.00 | 38.44 | 45.11 | 45.17 | 42.00 | 41.71 | 34.81 | 24.19* | **45.56** | 44.78* | 40.67 | 39.96 | 34.00 | 32.21* |
| UW | 87.03 | 89.59 | 92.42 | 92.41 | 91.23 | 91.19 | 71.29 | 90.07* | **94.43** | 93.21 | 91.31 | 90.78 | 71.11 | 70.68* |
| Average | 69.50 | 67.26 | 69.10 | 68.48 | 69.76 | 69.21 | 42.68 | 65.56 | 72.12 | 70.90 | **72.23** | 70.85 | 65.15 | 64.83 |

Table 2. Accuracy of all methods over 26 datasets. The best-performing approach is indicated by bold font.
\* = the difference between MimicShape and the corresponding base classifier is statistically significant (p<.05).

FCN and InceptionTime (IT). The Multivariate LSTM-FCN represents the state-of-the-art for a deep learning approach, and InceptionTime is a recent deep learning approach. These two methods both achieve high accuracy on multivariate time series classification problems. Also, we have added four additional classifiers to test Mimic's ability to simulate other classifiers. Our Mimic simulates the two deep learning classifiers and four different types of classifiers on all 26 datasets. Here we report accuracy performance based on ten-fold cross-validation. The results are summarized in Table 2.

As the table shows, in two cases the traditional shapelet-based classifiers outperform other classifiers and in one case the performance is identical. In the remaining 21 datasets, the other classifiers demonstrate the strongest performance. Thus, for most of these multivariate datasets the black-box classifiers perform better than shapelet-based classifiers. This motivates the need to create shapelets based on the black-box approach to balance recognition accuracy with interpretation. Also, from the table, we observe that the aggregated difference between MimicShape accuracy and accuracy of the emulated classifier across all datasets ranges from -2.5 to 0.1. The $p$-value for these differences is .4544; thus, the difference in performance is not significant at $p < .01$. Since the null hypotheses have not been rejected, we can regard the performance of MimicShape as statistically similar to that of the original classifier. In addition, we computed statistical significance of performance differences separately for each dataset. In this analysis, only the FD dataset yields a statistically different performance between the original classifier and MimicShape ($p<.05$). Also, the MimicShape version of the black-box classifiers performs better than the shapelet-based algorithm in 24 out of 26 datasets. These results indicate that we MimicShape offers a better-performing interpretable classifier for these 18 datasets.

We initially chose deep learners for our black-box approach because of their strong classification ability. However, we also want to determine whether MimicShape can Mimic other types of learning algorithms. For this reason, we also use MimicShape to emulate a decision tree classifier. Interestingly, MimicShape outperforms the decision tree classifier in 21 out of 26 datasets. The $p$-value is .000216; thus, the overall difference is performance is significant at $p<.05$. While the results demonstrate that MimicShape can be paired with multiple types of classifiers, for these datasets shapelet-based approaches as well as deep network approaches outperform decision tree classifiers.

Finally, we consider the importance map generated from the confidence score by masking features from the raw data. Since the decision tree can not create a distribution over the labels, the importance map will be vague for MimicShape to find the pattern. However, MimicShape can still find the pattern from the raw data. In this scenario, the MimicShape will become a classifier closer to DTW instead of decision tree.

## 5. Conclusion

Recently, more and more accurate models have been introduced for multivariate time series classification. The demand for interpretability and transparency are increasing, especially in some curial scenario. We proposed a novel algorithm, Mimic, which can adapt to the 'black-box' classifier to simulate the result and give a way to understand the 'black-box' classifier. Mimic is a pattern-based method that offers a path to understands the 'black-box' classifier from input data. Our unique contribution is that we provide a new perspective that instead of understanding complex 'black-box' classifiers, we understand the influence of input data for 'black-box' classifiers making decisions. From the experiment, we have shown Mimic can simulate the performance of the 'black-box' classifier. In the total 26 datasets, only one dataset shows a significant difference between the Mimic and target classifier. However, there are still some limitations to the current approach. Mimic has a hard time simulating the classifier that does not generate a distribution. The experiment shows, Mimic can poorly mimic the decision tree. For future work, we would like to extend the method to adapt the classifier that can not generate a distribution in future work. Also, we would like to try different approaches that mining the pattern for the input data.